\pdfoutput=1

\documentclass[11pt]{article}

\usepackage[]{ACL2023}

\usepackage{times}
\usepackage{latexsym}

\usepackage[T1]{fontenc}

\usepackage[utf8]{inputenc}

\usepackage{microtype}

\usepackage{inconsolata}
\usepackage{amsmath, bm}
\usepackage{graphicx}
\usepackage{amsfonts}
\usepackage{enumitem}

\usepackage{diagbox}
\usepackage{multirow}

\usepackage{csquotes}
\usepackage{hyperref}
\usepackage{comment}

\usepackage{array}
\usepackage{diagbox}
\usepackage{adjustbox}
\usepackage{verbatim}

\usepackage{graphicx}
\usepackage{lscape}

\usepackage{amssymb}
\usepackage{tikz}
\usepackage{soul}

\usepackage{xcolor}
\usepackage{ulem} 
\newcommand\highlight[1][yellow]{%
  \bgroup %
  \markoverwith{\textcolor{#1}{\vrule width.1em height.8em depth.2em}}%
  \ULon %
}

\title{Towards Effective Counter-Responses: Aligning Human Preferences with Strategies to Combat Online Trolling}

\author{Huije Lee\hspace{8mm}
    Hoyun Song\hspace{8mm}
    Jisu Shin\hspace{8mm}
    Sukmin Cho\\
    \textbf{SeungYoon Han}\hspace{8mm}
    \textbf{Jong C. Park}$\thanks{\hspace{2mm}Corresponding author}$ \\
    School of Computing \\
    Korea Advanced Institute of Science and Technology (KAIST)\\
    \texttt{\{huijelee,hysong,jisu.shin,nelllpic,seungyoonee,jongpark\}@kaist.ac.kr}\\
  }

\begin{document}
\maketitle

\begin{abstract}
Trolling in online communities typically involves disruptive behaviors such as provoking anger and manipulating discussions, leading to a polarized atmosphere and emotional distress.
Robust moderation is essential for mitigating these negative impacts and maintaining a healthy and constructive community atmosphere.
However, effectively addressing trolls is difficult because their behaviors vary widely and require different response strategies (RSs) to counter them.
This diversity makes it challenging to choose an appropriate RS for each specific situation.
To address this challenge, our research investigates whether humans have preferred strategies tailored to different types of trolling behaviors.
Our findings reveal a correlation between the types of trolling encountered and the preferred RS. 
In this paper, we introduce a methodology for generating counter-responses to trolls by recommending appropriate RSs, supported by a dataset aligning these strategies with human preferences across various troll contexts\footnote{Our dataset is publicly available at \url{https://github.com/huijelee/ELF-HumanPreference}.}.
The experimental results demonstrate that our proposed approach guides constructive discussion and reduces the negative effects of trolls, thereby enhancing the online community environment.
\end{abstract}

\section{Introduction}

In online communities, trolling is characterized as a disruptive activity, such as teasing, provoking anger, offending others, dominating discussions, or manipulating opinions~\cite{mihaylov-nakov-2016-hunting, golf-papez2017dont}.
Such behaviors often interfere with the productive exchange of ideas~\cite{bishop2013art}, contribute to polarized and hostile atmospheres~\cite{craker2016dark}, and cause significant emotional distress to victims~\cite{camacho2018cyberbullying}.
To preserve a positive community atmosphere, moderation is essential, as it helps mitigate the impact of trolling and maintain the continuity of constructive discussions~\cite{wise2006moderation,kraut2012building}.

However, determining the appropriate response to trolls is not straightforward.
As \citet{hardaker2010trolling} noted, the range of trolling behaviors is diverse, and the corresponding response strategies for addressing them should vary accordingly.
For example, when faced with highly politicized and offensive comments, responses should explicitly and strongly incorporate clear warnings.
By contrast, when encountering off-topic opinions during focused discussions, responses should gently guide them to realign their contributions with the goals of the discussion.
This range of behaviors and required responses adds to the challenge of choosing the most appropriate strategy for each specific situation.

A recent study~\cite{mun-etal-2023-beyond} has found that humans tend to prefer certain strategies when countering hate speech.
Inspired by this finding, we hypothesized that humans might also have a preferred response tailored to each distinct troll situation.
To investigate this, we explored whether preferences exist for various response strategies to different trolling behaviors.
Our findings showed a clear correlation between the types of trolling encountered and response strategies preferred, enhancing our understanding of how to counter different trolling behaviors appropriately.

In this paper, we aim to develop a method for generating the most effective strategy for responding to trolls in diverse situations, thereby promoting a desirable online community environment.
Accordingly, we propose a method that recommends a specific response strategy for each type of trolling behavior, which enables the generation of appropriate \textbf{C}ounter-\textbf{R}esponses (CR) to trolls aligned with human preference.
To this end, we investigated the relationship between different \textbf{T}rolling \textbf{S}trategies (TS) and the corresponding preferred \textbf{R}esponse \textbf{S}trategies (RS).
Then, we constructed a dataset that matches RS to user preferences across various troll contexts.
Utilizing this dataset, we developed a recommendation system for RS and designed a CR generation methodology that selects the most appropriate strategy based on this system.
Our experimental results demonstrate the gap between CRs generated by general-purpose Large Language Models (LLMs) and human-preferable CRs, highlighting the importance of aligning human preferences with strategies for effective CR generation.

Our contributions and findings are threefold:
\begin{itemize}[leftmargin=*,topsep=-2px,partopsep=0px]
\setlength\itemsep{-0.5em}
    \item This is the first study to explore the relationship between human preferences and response strategies for addressing various trolling behaviors, shedding light on novel approaches for managing online communities.
    \item We propose a novel CR generation methodology, aligning user preferences with response strategies, and enhancing the effectiveness of automatic moderation.
    \item Our experimental results demonstrate that our proposed approach guides constructive discussion and mitigates the negative impacts of trolls.
\end{itemize}

\section{Related Works}
Trolling behaviors vary widely, from explicit expressions of hate, such as promoting discrimination based on gender, to subtle annoyance, including digressing onto irrelevant topics or misleading others with harmful advice~\cite{herring2002searching, hardaker2010trolling,fichman-sanfilippo-2016-online,mihaylov-nakov-2016-hunting,bratu2017inexorable,golf-papez2017dont}.
\citet{hardaker2013uh} outlined the types of trolling strategies ranging from covert to overt and examined the types of response strategies accordingly.
Attempts to implement automatic counter-trolling have been made~\cite{chung-etal-2021-towards,zhu-bhat-2021-generate,ELF,gupta-etal-2023-counterspeeches,furman-etal-2023-high,yu2023fine}, but the challenge of automatically selecting the appropriate RS still remains.
In this study, we explore effective CR generation strategies to address these gaps.

When moderating trolls to preserve a healthy online community environment, a critical factor is community approval of the intervention approach~\cite{weld2022makes}.
Common responses to trolling include ignoring~\cite{li2023ignoring}, deleting comments~\cite{cheng2015antisocial,park-etal-2021-detecting-community}, and banning users or communities~\cite{chandrasekharan2017you}.
However, these approaches have been criticized for potential contagion of such behavior~\cite{cheng2017anyone}, leading to censorship accusations~\cite{richards2000counterspeech}, and neglecting user feedback~\cite{myers2018censored}. While recent advancements in LLMs have led to instruction-integrated interactive moderation~\cite{zheng-etal-2023-makes,cho-etal-2024-language} showing impressive response generation capabilities, there remains a need for more targeted approaches to combat trolling effectively, as \citet{zheng-etal-2023-makes} found that the commonly used gentle guiding approach is not universally preferred. 
In this paper, we explore how to choose the appropriate RS for countering trolls, motivated by the previous research that highlights significant variations in preferences for responding to hate speech~\cite{mun-etal-2023-beyond}.

\section{Methodology}
\label{section:method}

In this section, we explore the relationship between TS and preferred RS, detailing the process we used to construct a dataset that aligns human preferences with RS. 
Our dataset comprises troll comments paired with CRs preferred by human participants, selected from multiple CRs.
Furthermore, we outline our method for generating CRs by leveraging the distribution of RS derived from this dataset.

\subsection{Data Collection}

Our data collection involves crawling posts and troll comments from various subreddits on Reddit published in 2022.
To ensure that collected posts and comments provide adequate contextual information for understanding discussions, we applied a character limit of a minimum of 12 and a maximum of 512 characters. We excluded texts deleted by Reddit or users and samples containing external links or media materials to prevent loss of contextual information due to embedded links, photos, or videos. To gather potential troll comments, we first selected posts containing root downvoted comments. We then employed instruction-tuned GPT-3.5~\cite{openai-2022-chatgpt} for troll classification. Further details for the troll classification are shown in Appendix~\ref{appendix:troll_classifier}.

\subsection{Data Annotation}    
\label{section:data_annotation}

We adopted the taxonomy of trolling behavior developed by \citet{hardaker2013uh}, which classifies TS ranging from covert to overt. This taxonomy classifies trolling behaviors along a continuum, starting from the covert strategy, such as \textit{Digression}, to the overt strategy, \textit{Aggression}. For categorizing counter-responses, we utilized a set of seven response strategies~\cite{hardaker2015i}. These strategies include \textit{Engage}, \textit{Ignore}, and \textit{Expose} as nudging responses, and \textit{Challenge}, \textit{Critique}, \textit{Mock}, and \textit{Reciprocate} as confrontational responses. Detailed descriptions of TS and RS are provided in Appendix~\ref{appendix:trolling_and_response_strategies}.

We recruited six annotators and provided them with guidelines on both TS and RS. Annotators were given context information including the subreddit name, post title, and body text, along with a troll comment and seven model-generated counter-responses with different response strategies. For each sample, annotators labeled the perceived TS and selected the most preferable counter-response that resonates with, changes, or represents their views.

We conducted an offline QA session using the same 40 samples to ensure that they fully engaged and understood the annotation task. Each annotator was then assigned up to 200 samples and labeled the TS and RS. The annotators were instructed to skip samples that were unclear, had non-English content, and were not related to trolling. Finally, we collected a dataset of 875 labeled samples. Details for the annotation process are provided in Appendix~\ref{appendix b.1: annotation in rq1}.

\subsection{Investigation of Human Preference}
\begin{figure}[t!]
    \centering
    \includegraphics[width=\linewidth]{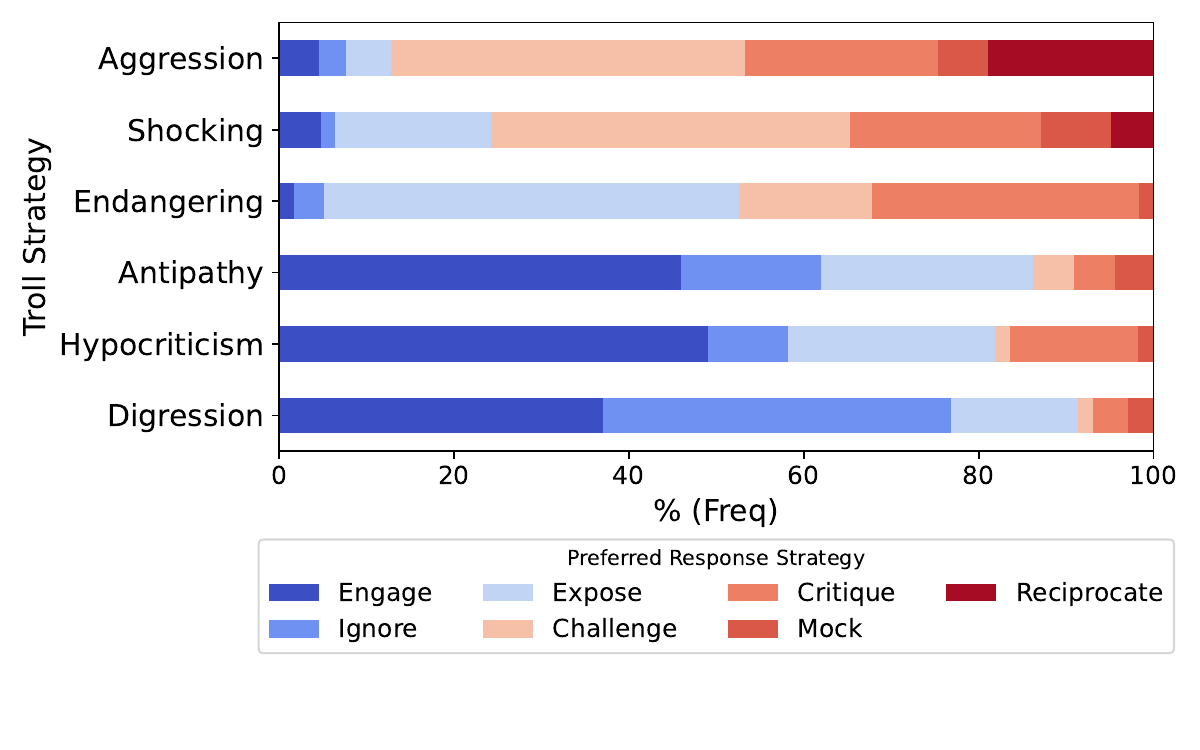}
    \vspace{-0.5in}
    \caption{Distribution of preferred RS relative to the TS. The top three bars indicate overt trolls, and the bottom three bars indicate covert trolls.}
    \label{fig: tsxrs}
    \vspace{-0.2in}
\end{figure}

Figure~\ref{fig: tsxrs} shows the distribution of preferred RS relative to the types of TS within our dataset. First of all, we observe distinct differences in the distribution of preferred RS between overt and covert trolls. Delving into the details of TS, we also observe a gradual increase in the preference for nudging strategies such as \textit{Engage}, \textit{Ignore}, and \textit{Expose} as moving from the most overt troll strategy, \textit{Aggression}, to the most covert troll strategy, \textit{Digression}. For overt trolls, \textit{Challenge} and \textit{Critique} strategies were predominantly preferred, while for covert trolls, \textit{Engage} and \textit{Expose} strategies were more favored. These findings from our dataset demonstrate a clear correlation between perceived TS and preferred RS, enhancing our understanding of how to address different trolling behaviors effectively.

\subsection{Counter-Response Generation}

Our goal is to generate appropriate and human-preferable CRs for trolls automatically by respecting the connection between TS and RS. 
Appropriateness, which we addressed, refers to the ability to protect a community by mitigating the influence of trolls and sustaining discussion in the community. Although LLMs can generate CR with human-like fluency, they are not yet fully able to produce appropriate and human-preferable responses~\cite{zheng-etal-2023-makes}. 

We propose a CR generation model guided by a Human-\textbf{P}referable \textbf{R}esponse \textbf{S}trategy (PRS).
Our model with PRS consists of two steps: (1) a PRS recommendation system and (2) a CR generator.
A \textbf{PRS recommendation system} takes a post, a troll comment, and the comment's TS as inputs and predicts which RS is preferred the most.
Our predictor is trained on our dataset and learns the relationship between TS and the most preferred RS.
Our \textbf{CR generator} takes the same input as the PRS recommendation system, along with the predicted PRS as an input, to generate CRs.
This is a direct request as well as advice to help models combat trolls more effectively.
Our CR generation model is expected to generate highly favorable responses by aligning closely with human preferences.

\section{Experiments}
\label{section: RQ3}

\begin{figure*}[t!]
    \centering
    \begin{minipage}{\textwidth}
        \centering
        \includegraphics[width=0.37\linewidth]{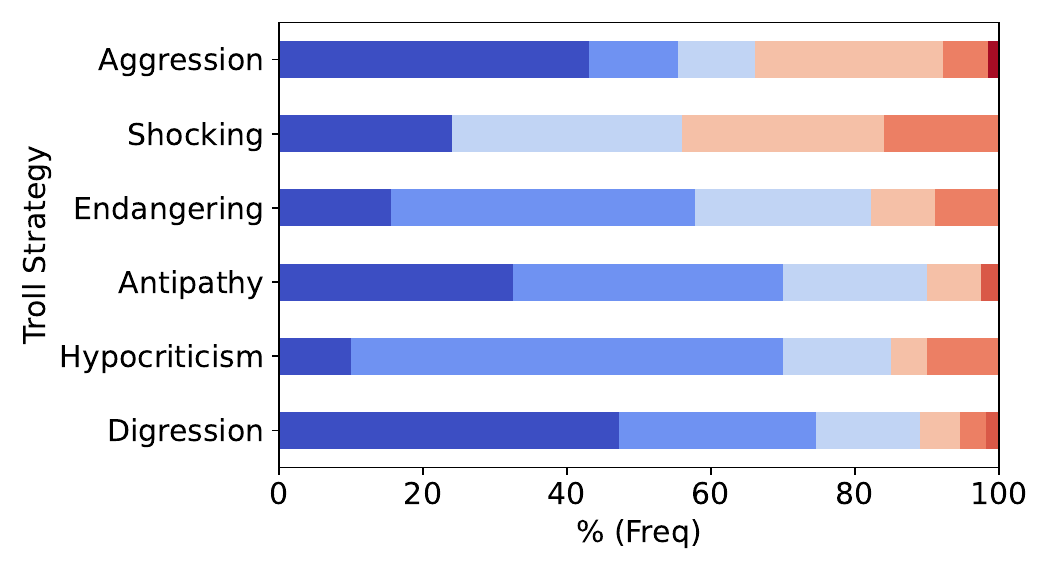}
        \hfill
        \hspace{-0.5in}
        \includegraphics[width=0.32\linewidth]{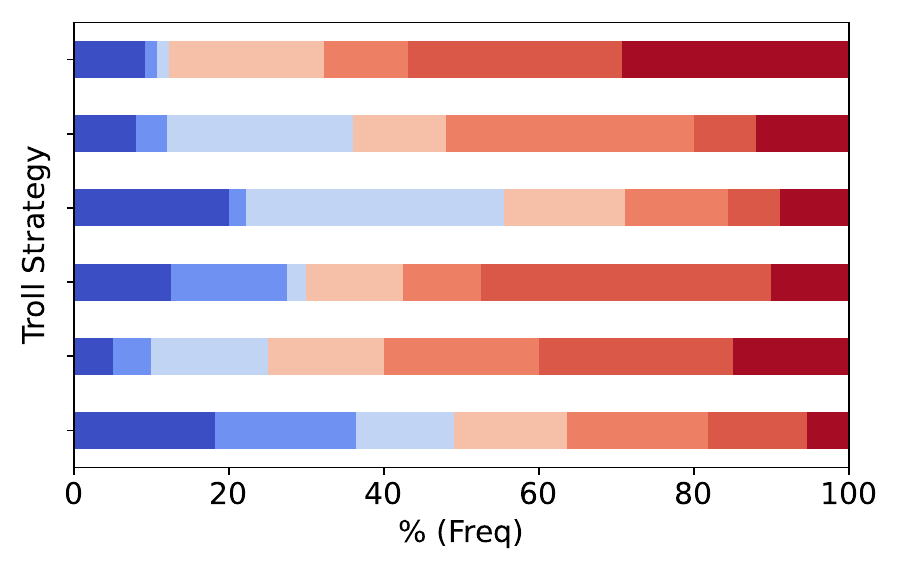}
        \hfill
        \hspace{-0.5in}
        \includegraphics[width=0.32\linewidth]{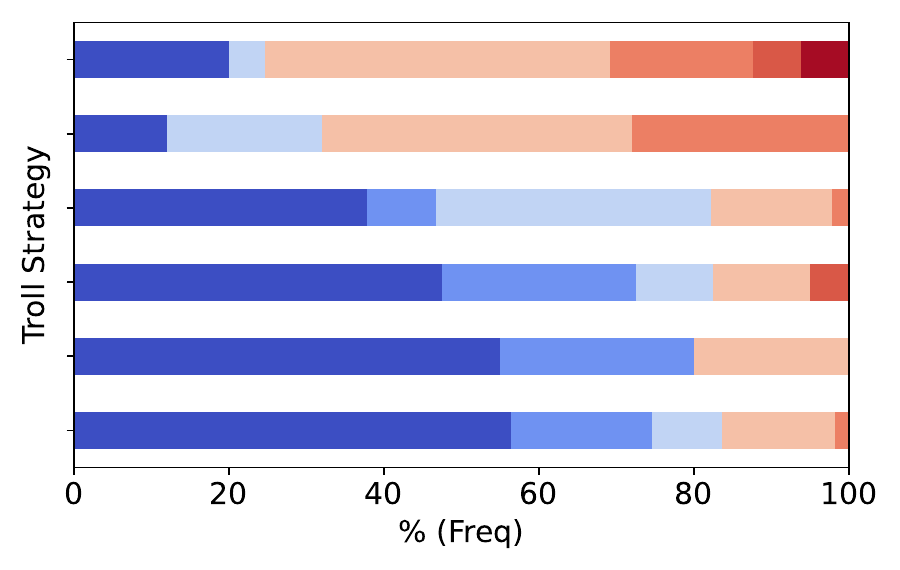}
    \end{minipage}

    \vspace{-0.08in}
    
    \includegraphics[width=0.9\textwidth]{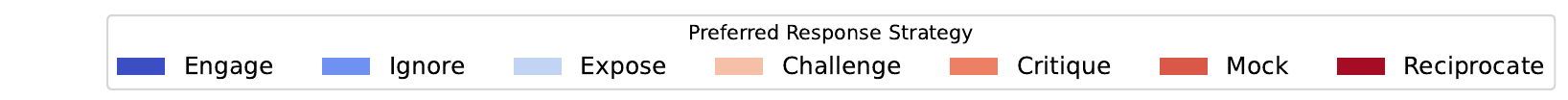}

    \vspace{-0.1in}

    \caption{Distribution of humans' perceived response strategies of generated responses (left: Default, center: Strategy-Provided, right: PRS (Ours)).}
    \label{fig:ratio_generated}

    \vspace{-0.2in}
    
\end{figure*}

In this section, we conduct experiments to evaluate the effectiveness of our proposed approach for generating CRs to trolls. To demonstrate the importance of aligning CRs with human preferences, we compare CRs produced by our model against those generated by existing models using human evaluation metrics.

\subsection{Experimental Setup}

\paragraph{Models}
We use GPT-3.5~\cite{openai-2022-chatgpt}, the accessible LLM capable of generating human-like sentences, as our default CR generator.
In our experiments, we compare three models in our experiments:
(1) \textbf{Default} model deals only with an online post and a troll comment left on the post for its generation.
(2) \textbf{Strategy-Provided (SP)} model is instructed with definitions of TS and RS, along with in-context examples for each RS. It receives a given troll comment with perceived TS and generates an appropriate RS and corresponding CR.
(3) Our model (\textbf{PRS}) performs under the same settings as SP, but it additionally receives the predicted PRS and in-context examples tailored to this PRS. For the PRS recommendation system, we fine-tuned Flan-T5 Large~\cite{Flan-T5}. Details of the experimental setup are provided in Appendix~\ref{appendix: experiment}.

\paragraph{Test Dataset}
We additionally collect 50 troll comments and annotate them in the same manner described in Section~\ref{section:data_annotation}.

\paragraph{Evaluation Metrics}

To evaluate the effectiveness of CRs, we focus on their impact to promote constructive discussions and mitigate the negative impacts of trolling, rather than attempting to measure the persuasion of trolls. Troll users often view any attention as `mission accomplished'\cite{golf-papez2017dont}, making it challenging to assess the direct impact on their behavior. Instead, we designed our evaluation process to directly ask evaluators to assess CRs from the perspective of general Reddit users.

We recruited five evaluators to assess the generated responses in the test dataset across three key aspects:
1)~\textbf{Preference} assesses how well the responses resonate with, change, or represent their views. Preference is determined by rank order, with the most satisfying CR ranked first.
2)~\textbf{Constructiveness} measures how effectively a counter-response maintains focus on the topic and creates a welcoming environment that encourages broader participation in the discussion.
A high constructiveness score indicates that the response has facilitated constructive discussion and encouraged participation, whereas a low score suggests that it has escalated conflict or derailed the conversation.
3)~\textbf{Supportiveness} evaluates how well a counter-response defends targeted individuals or groups, supporting them against negative effects of trolls.
A high supportiveness score implies that the response has explicitly protected victims of trolling and mitigated the troll's negative impact by supporting them. Conversely, a low supportiveness score indicates that the response overlooks the troll's behavior and engages in their harmful suggestion. These two criteria are measured on a 5-point Likert scale.
Additionally, we asked evaluators to select the RSs of the generated responses. More details of the annotation scheme are presented in Appendix~\ref{appendix:evaluation_scheme}.

\begin{figure}[t!]
    \centering
    \includegraphics[width=0.9\linewidth]{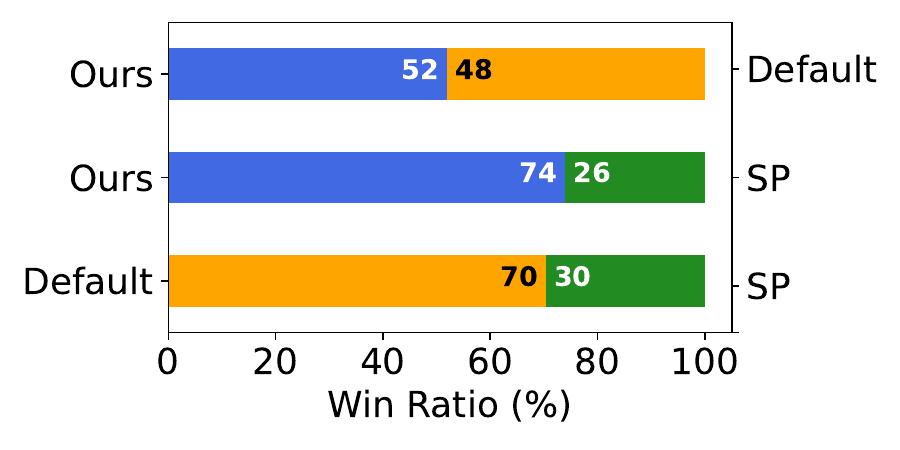}
    \vspace{-0.2in}
    \caption{Visualization of the rank test for preference.}
    \label{fig:AB_test}
    \vspace{-0.2in}
\end{figure}
\begin{figure}[t!]
    \centering
    \begin{minipage}{\columnwidth}
        \includegraphics[width=0.5\linewidth]{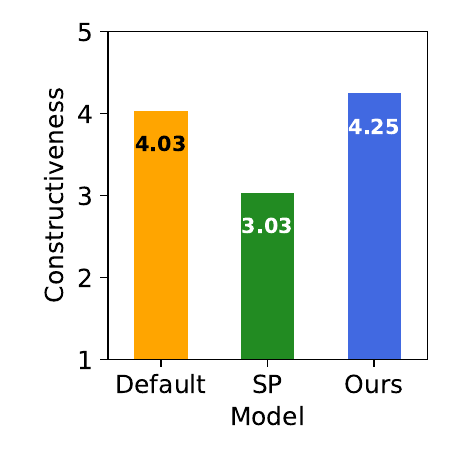}
        \hfill
        \hspace{-0.3in}
        \includegraphics[width=0.5\linewidth]{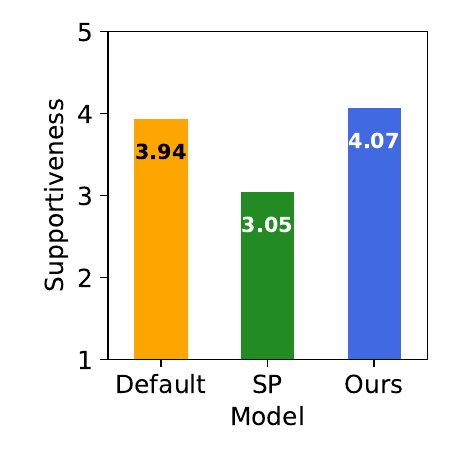}
    \end{minipage}
    \vspace{-0.2in}
    \caption{The result scores of our experiments (left: Constructiveness, right: Supportiveness).}
    \vspace{-0.15in}
    \label{fig:likert_bar}
\end{figure}

\subsection{Experimental Results and Discussions}

\paragraph{Preference}
Figure~\ref{fig:AB_test} presents the win ratios of AB testing that we converted the preference rankings of three methods.
Default and our model beat the Strategy-Provided model by over 70\%, and ours beats the Default model by a small margin (4.8\%p).
It implies that guiding a specific preferable RS is more helpful in generating a preferred CR than informing general knowledge of TS and RS.
We display the distribution of humans' perceived RS of generated responses in Figure~\ref{fig:ratio_generated}.
The Default model generally responded using \textit{Nudging} strategies, while the Strategy-Provided model utilized \textit{Confrontational} strategies against most trolls.
However, our model used flexible RS: the \textit{Confrontational} strategies to overt trolls and \textit{Nudging} strategies to covert trolls.

\begin{table}[t!]
\centering
\small
\begin{tabular}{lcccc}
\hline
\multirow{2}{*}{\textbf{Model}} & \multicolumn{2}{c}{\textbf{Coarse-grained}} & \multicolumn{2}{c}{\textbf{Fine-grained}} \\
 & \textbf{JSD ($\downarrow$)} & \textbf{HD ($\downarrow$)} & \textbf{JSD ($\downarrow$)} & \textbf{HD ($\downarrow$)} \\
\hline
Default & 0.253 & 0.257 & 0.378 & 0.404 \\
SP & 0.288 & 0.292 & 0.409 & 0.433 \\
\textbf{Ours} & \textbf{0.156} & \textbf{0.157} & \textbf{0.338} & \textbf{0.365} \\
\hline
\end{tabular}
\caption{Comparison of distributional similarity between model-generated and human-preferred strategies. Coarse-grained refers to overt/covert and nudging/confrontational categories, while fine-grained refers to detailed TS and RS categories.}
\label{tab:jsd_hd_comparison}
\end{table}

Comparing the distributions of generated RS in Figure~\ref{fig:ratio_generated} and human preference in Figure~\ref{fig: tsxrs}, our model succeeded in forming the distribution that most closely resembles that of human-preferred RS.
To quantify the alignment between generated responses and human preferences, we calculated the Jensen-Shannon Distance (JSD) and Hellinger Distance (HD) for each model. Table~\ref{tab:jsd_hd_comparison} shows the results. Our model achieved the lowest JSD and HD in both granularities, indicating the closest alignment with human preferences. The Strategy-Provided model showed the poorest alignment, suggesting that merely providing strategy information without considering context-specific human preferences may lead to suboptimal responses. Detailed explanations of JSD and HD calculations are provided in Appendix~\ref{appendix:jsd_hd_metrics}.

\paragraph{Constructiveness}

Our model achieved the highest constructiveness score of 4.25 compared to the baseline scores of 4.03 for the Default and 3.03 for the Strategy-Provided model (see Figure~\ref{fig:likert_bar}).
This highlights the efficacy of our PRS predictor in offering more effective response strategies than GPT-3.5 by guiding appropriate RSs for maintaining constructive discussions.
In practical cases, our model improved discussion quality by generating responses that indicated off-topic comments from trolls and reminded the original topic to refocus on the conversation.

\paragraph{Supportiveness}
Our model achieved the highest supportiveness at 4.07, compared to 3.94 for the Default and 3.05 for the Strategy-Provided model. 
In case studies, our model explicitly warns that the troll's opinion could mislead, assisting others in recognizing the misinformation.
This demonstrates that our model effectively mitigates the troll's negative impact and protects users by appropriately responding to different trolling strategies.
We provide details of the significance tests and case studies in Appendix~\ref{appendix: experiment}.

\section{Conclusion}

In this work, we explored the importance of aligning human preferences with response strategies to address the challenge of trolling in online communities.
We constructed a dataset via human annotation that captures the relationship between types of trolling and preferred counter-responses. This dataset showed distinct differences in preferences for response strategies depending on various troll strategies.
In our experiments, we leveraged this relationship to generate human-preferred and effective counter-responses. Our approach not only succeeded in generating more preferred counter-responses but also promoted constructive discussions and mitigated the harmful impact of trolling.

For future work, we recommend a deeper analysis of preference differences across demographics and communities to further enhance online community health and user engagement. This research paves the way for further advancements in interactive moderation, enabling more targeted and effective approaches to combat trolling.

\section*{Limitations}

In this study, we collected about 900 labeled data.
The limited size of the dataset is due to the exclusion of non-troll comments from the initially crawled datasets. Additionally, constraints such as budget limitations, the limited availability of annotators, and annotator fatigue restricted our capacity to label a larger dataset. These limitations also prevented us from applying a variety of training approaches, such as supervised fine-tuning (SFT)~\cite{tekiroglu-etal-2020-generating, chung-etal-2021-towards, ELF} or reinforcement learning from human feedback (RLHF)~\cite{ouyang2022training} with the PPO algorithm~\cite{PPO}, with Large Language Models (LLMs) like LLaMA~\cite{touvron2023llama2} and Mixtral~\cite{jiang2024mixtral}. Therefore, we adopted a methodology utilizing an accessible LLM, GPT-3.5, with in-context learning. Despite its size, however, our dataset reveals clear patterns between troll strategies and response strategies. As the experiment expands and more data is collected, we expect that our methodology can be utilized in various ways. This aspect falls outside the scope of our current research and will be addressed in future work.

Although we provide the annotators with detailed guidelines to facilitate a clear understanding of troll strategies and response strategies, there are still differences in perceptions of trolling and preferences of counter-response.
Also, as the dataset has been annotated with trolling strategies, response strategies, and human preferences from the perspective of general Reddit users, variations in annotations may arise due to differences in the annotators' understanding of the context and culture of specific communities.
Perceived trolling points, which are linked to community understanding, can vary and thus influence the choices of preferred response strategies. However, these differences also mirror real-world variations~\cite{weld2022makes} and can be viewed as a natural diversity of opinions.

Our proposed approach, which generates appropriate responses to perceived trolls, can be utilized alongside judgments on trolling that may involve automated decisions using user flagging or moderator determinations. This enables its application as an automatic counter-response generation system. While automatic counter-response generation systems avoid the problem of censorship, they can still manifest biases and result in unintended consequences \cite{ferrara2023should}. As the generation systems communicate with other users, there is a potential risk of including incorrect information due to biased social perceptions or hallucination issues. Despite these risks, we believe that further investigation and analysis of these systems could provide valuable insights and guidance on how online communities can adapt, practice, and moderate in an era filled with AI-generated content~\cite{lloyd2023there,zhao2024adapting}.

\section*{Ethics Statement}

Our annotation experiment was approved by the Institutional Review Board (IRB)\footnote{Approval number: KH2023-166}. All participants in annotation tasks indicated their understanding of the procedure for the annotation and acknowledged their agreement to participate. 
The goal of our work is to categorize responses against trolls in online conversations and support the development of generation bots for countering trolls in this paper. Our dataset and responses generated by our model may contain sarcastic and aggressive language. We tried to observe how they communicate as-is, even though it could include socially biased content or hate speech.

\section*{Acknowledgements}
This work was supported by Institute for Information and communications Technology Promotion (IITP) grant funded by the Korea government (No. 2018-0-00582, Prediction and augmentation of the credibility distribution via linguistic analysis and automated evidence document collection) and the Artificial intelligence industrial convergence cluster development project funded by the Ministry of Science and ICT (MSIT, Korea) \& Gwangju Metropolitan City.


\normalem
\bibliography{custom}

\begin{thebibliography}{46}
\expandafter\ifx\csname natexlab\endcsname\relax\def\natexlab#1{#1}\fi

\bibitem[{Beran(1977)}]{beran1977minimum}
Rudolf Beran. 1977.
\newblock Minimum hellinger distance estimates for parametric models.
\newblock \emph{The annals of Statistics}, pages 445--463.

\bibitem[{Bishop(2013)}]{bishop2013art}
Jonathan Bishop. 2013.
\newblock The art of trolling law enforcement: a review and model for implementing ‘flame trolling'legislation enacted in great britain (1981--2012).
\newblock \emph{International Review of Law, Computers \& Technology}, 27(3):301--318.

\bibitem[{Bratu(2017)}]{bratu2017inexorable}
Sofia Bratu. 2017.
\newblock The inexorable shift towards an increasingly hostile cyberspace environment: The adverse social impact of online trolling behavior.
\newblock \emph{Contemporary Readings in Law and Social Justice}, 9(2):88--94.

\bibitem[{Brown et~al.(2020)Brown, Mann, Ryder, Subbiah, Kaplan, Dhariwal, Neelakantan, Shyam, Sastry, Askell et~al.}]{brown2020language}
Tom Brown, Benjamin Mann, Nick Ryder, Melanie Subbiah, Jared~D Kaplan, Prafulla Dhariwal, Arvind Neelakantan, Pranav Shyam, Girish Sastry, Amanda Askell, et~al. 2020.
\newblock Language models are few-shot learners.
\newblock \emph{Advances in neural information processing systems}, 33:1877--1901.

\bibitem[{Camacho et~al.(2018)Camacho, Hassanein, and Head}]{camacho2018cyberbullying}
Sonia Camacho, Khaled Hassanein, and Milena Head. 2018.
\newblock Cyberbullying impacts on victims’ satisfaction with information and communication technologies: The role of perceived cyberbullying severity.
\newblock \emph{Information \& Management}, 55(4):494--507.

\bibitem[{Chandrasekharan et~al.(2017)Chandrasekharan, Pavalanathan, Srinivasan, Glynn, Eisenstein, and Gilbert}]{chandrasekharan2017you}
Eshwar Chandrasekharan, Umashanthi Pavalanathan, Anirudh Srinivasan, Adam Glynn, Jacob Eisenstein, and Eric Gilbert. 2017.
\newblock You can't stay here: The efficacy of reddit's 2015 ban examined through hate speech.
\newblock \emph{Proceedings of the ACM on human-computer interaction}, 1(CSCW):1--22.

\bibitem[{Cheng et~al.(2017)Cheng, Bernstein, Danescu-Niculescu-Mizil, and Leskovec}]{cheng2017anyone}
Justin Cheng, Michael Bernstein, Cristian Danescu-Niculescu-Mizil, and Jure Leskovec. 2017.
\newblock Anyone can become a troll: Causes of trolling behavior in online discussions.
\newblock In \emph{Proceedings of the 2017 ACM conference on computer supported cooperative work and social computing}, pages 1217--1230.

\bibitem[{Cheng et~al.(2015)Cheng, Danescu-Niculescu-Mizil, and Leskovec}]{cheng2015antisocial}
Justin Cheng, Cristian Danescu-Niculescu-Mizil, and Jure Leskovec. 2015.
\newblock Antisocial behavior in online discussion communities.
\newblock In \emph{Proceedings of the international aaai conference on web and social media}, volume~9, pages 61--70.

\bibitem[{Cho et~al.(2024)Cho, Liu, Shi, Jain, Rizk, Huang, Lu, Wen, Gratch, Ferrara, and May}]{cho-etal-2024-language}
Hyundong Cho, Shuai Liu, Taiwei Shi, Darpan Jain, Basem Rizk, Yuyang Huang, Zixun Lu, Nuan Wen, Jonathan Gratch, Emilio Ferrara, and Jonathan May. 2024.
\newblock Can language model moderators improve the health of online discourse?
\newblock In \emph{Proceedings of the 2024 Conference of the North American Chapter of the Association for Computational Linguistics: Human Language Technologies (Volume 1: Long Papers)}, pages 7478--7496.

\bibitem[{Chung et~al.(2022)Chung, Hou, Longpre, Zoph, Tay, Fedus, Li, Wang, Dehghani, Brahma, Webson, Gu, Dai, Suzgun, Chen, Chowdhery, Narang, Mishra, Yu, Zhao, Huang, Dai, Yu, Petrov, Chi, Dean, Devlin, Roberts, Zhou, Le, and Wei}]{Flan-T5}
Hyung~Won Chung, Le~Hou, Shayne Longpre, Barret Zoph, Yi~Tay, William Fedus, Eric Li, Xuezhi Wang, Mostafa Dehghani, Siddhartha Brahma, Albert Webson, Shixiang~Shane Gu, Zhuyun Dai, Mirac Suzgun, Xinyun Chen, Aakanksha Chowdhery, Sharan Narang, Gaurav Mishra, Adams Yu, Vincent~Y. Zhao, Yanping Huang, Andrew~M. Dai, Hongkun Yu, Slav Petrov, Ed~H. Chi, Jeff Dean, Jacob Devlin, Adam Roberts, Denny Zhou, Quoc~V. Le, and Jason Wei. 2022.
\newblock \href {https://doi.org/10.48550/ARXIV.2210.11416} {Scaling instruction-finetuned language models}.
\newblock \emph{CoRR}, abs/2210.11416.

\bibitem[{Chung et~al.(2021)Chung, Tekiro{\u{g}}lu, and Guerini}]{chung-etal-2021-towards}
Yi-Ling Chung, Serra~Sinem Tekiro{\u{g}}lu, and Marco Guerini. 2021.
\newblock \href {https://doi.org/10.18653/v1/2021.findings-acl.79} {Towards knowledge-grounded counter narrative generation for hate speech}.
\newblock In \emph{Findings of the Association for Computational Linguistics: ACL-IJCNLP 2021}. Association for Computational Linguistics.

\bibitem[{Craker and March(2016)}]{craker2016dark}
Naomi Craker and Evita March. 2016.
\newblock The dark side of facebook{\textregistered}: The dark tetrad, negative social potency, and trolling behaviours.
\newblock \emph{Personality and Individual Differences}, 102:79--84.

\bibitem[{Endres and Schindelin(2003)}]{endres2003new}
Dominik~Maria Endres and Johannes~E Schindelin. 2003.
\newblock A new metric for probability distributions.
\newblock \emph{IEEE Transactions on Information theory}, 49(7):1858--1860.

\bibitem[{Ferrara(2023)}]{ferrara2023should}
Emilio Ferrara. 2023.
\newblock Should chatgpt be biased? challenges and risks of bias in large language models.
\newblock \emph{arXiv preprint arXiv:2304.03738}.

\bibitem[{Fichman and Sanfilippo(2016)}]{fichman-sanfilippo-2016-online}
Pnina Fichman and Madelyn~R. Sanfilippo. 2016.
\newblock \emph{Online Trolling and Its Perpetrators: Under the Cyberbridge}.
\newblock Rowman \& Littlefield Publishers, Inc.

\bibitem[{Furman et~al.(2023)Furman, Torres, Rodr{\'\i}guez, Letzen, Martinez, and Alemany}]{furman-etal-2023-high}
Dami{\'a}n Furman, Pablo Torres, Jos{\'e} Rodr{\'\i}guez, Diego Letzen, Maria Martinez, and Laura Alemany. 2023.
\newblock High-quality argumentative information in low resources approaches improve counter-narrative generation.
\newblock In \emph{Findings of the Association for Computational Linguistics: EMNLP 2023}, pages 2942--2956.

\bibitem[{Golf-Papez and Veer(2017)}]{golf-papez2017dont}
Maja Golf-Papez and Ekant Veer. 2017.
\newblock \href {https://doi.org/10.1080/0267257X.2017.1383298} {Don’t feed the trolling: rethinking how online trolling is being defined and combated}.
\newblock \emph{Journal of Marketing Management}, 33(15-16):1336--1354.

\bibitem[{Gupta et~al.(2023)Gupta, Desai, Goel, Bandhakavi, Chakraborty, and Akhtar}]{gupta-etal-2023-counterspeeches}
Rishabh Gupta, Shaily Desai, Manvi Goel, Anil Bandhakavi, Tanmoy Chakraborty, and Md.~Shad Akhtar. 2023.
\newblock Counterspeeches up my sleeve! intent distribution learning and persistent fusion for intent-conditioned counterspeech generation.
\newblock In \emph{Proceedings of the 61st Annual Meeting of the Association for Computational Linguistics (Volume 1: Long Papers)}, pages 5792--5809.

\bibitem[{Hardaker(2010)}]{hardaker2010trolling}
Claire Hardaker. 2010.
\newblock Trolling in asynchronous computer-mediated communication: From user discussions to academic definitions.

\bibitem[{Hardaker(2013)}]{hardaker2013uh}
Claire Hardaker. 2013.
\newblock \href {https://doi.org/10.1075/jlac.1.1.04har} {“uh. . . . not to be nitpicky,,,,,but…the past tense of drag is dragged, not drug.”: An overview of trolling strategies}.
\newblock \emph{Journal of Language Aggression and Conflict}, 1:58--86.

\bibitem[{Hardaker(2015)}]{hardaker2015i}
Claire Hardaker. 2015.
\newblock \href {https://doi.org/10.3366/cor.2015.0074} {‘i refuse to respond to this obvious troll’: an overview of responses to (perceived) trolling}.
\newblock \emph{Corpora}, 10(2):201--229.

\bibitem[{Herring et~al.(2002)Herring, Job-Sluder, Scheckler, and Barab}]{herring2002searching}
Susan Herring, Kirk Job-Sluder, Rebecca Scheckler, and Sasha Barab. 2002.
\newblock Searching for safety online: Managing" trolling" in a feminist forum.
\newblock \emph{The information society}, 18(5):371--384.

\bibitem[{Jiang et~al.(2024)Jiang, Sablayrolles, Roux, Mensch, Savary, Bamford, Chaplot, Casas, Hanna, Bressand et~al.}]{jiang2024mixtral}
Albert~Q Jiang, Alexandre Sablayrolles, Antoine Roux, Arthur Mensch, Blanche Savary, Chris Bamford, Devendra~Singh Chaplot, Diego de~las Casas, Emma~Bou Hanna, Florian Bressand, et~al. 2024.
\newblock Mixtral of experts.
\newblock \emph{arXiv preprint arXiv:2401.04088}.

\bibitem[{Kraut and Resnick(2012)}]{kraut2012building}
Robert~E Kraut and Paul Resnick. 2012.
\newblock \emph{Building successful online communities: Evidence-based social design}.
\newblock Mit Press.

\bibitem[{Lee et~al.(2022)Lee, Na, Song, Shin, and Park}]{ELF}
Huije Lee, Young~Ju Na, Hoyun Song, Jisu Shin, and Jong~C. Park. 2022.
\newblock \href {https://aclanthology.org/2022.lrec-1.378} {{ELF22:} {A} context-based counter trolling dataset to combat internet trolls}.
\newblock In \emph{Proceedings of the Thirteenth Language Resources and Evaluation Conference, {LREC} 2022, Marseille, France, 20-25 June 2022}, pages 3530--3541. European Language Resources Association.

\bibitem[{Li et~al.(2023)Li, Cai, and Wohn}]{li2023ignoring}
Na~Li, Jie Cai, and Donghee~Yvette Wohn. 2023.
\newblock Ignoring as a moderation strategy for volunteer moderators on twitch.
\newblock In \emph{Extended Abstracts of the 2023 CHI Conference on Human Factors in Computing Systems}, pages 1--7.

\bibitem[{Lloyd et~al.(2023)Lloyd, Reagle, and Naaman}]{lloyd2023there}
Travis Lloyd, Joseph Reagle, and Mor Naaman. 2023.
\newblock "there has to be a lot that we're missing": Moderating ai-generated content on reddit.
\newblock \emph{arXiv preprint arXiv:2311.12702}.

\bibitem[{Loshchilov and Hutter(2017)}]{loshchilov2017decoupled}
Ilya Loshchilov and Frank Hutter. 2017.
\newblock Decoupled weight decay regularization.
\newblock \emph{arXiv preprint arXiv:1711.05101}.

\bibitem[{Mihaylov and Nakov(2016)}]{mihaylov-nakov-2016-hunting}
Todor Mihaylov and Preslav Nakov. 2016.
\newblock \href {https://doi.org/10.18653/v1/P16-2065} {Hunting for troll comments in news community forums}.
\newblock In \emph{Proceedings of the 54th Annual Meeting of the Association for Computational Linguistics (Volume 2: Short Papers)}, pages 399--405, Berlin, Germany. Association for Computational Linguistics.

\bibitem[{Min et~al.(2022)Min, Lyu, Holtzman, Artetxe, Lewis, Hajishirzi, and Zettlemoyer}]{min-etal-2022-rethinking}
Sewon Min, Xinxi Lyu, Ari Holtzman, Mikel Artetxe, Mike Lewis, Hannaneh Hajishirzi, and Luke Zettlemoyer. 2022.
\newblock \href {https://doi.org/10.18653/v1/2022.emnlp-main.759} {Rethinking the role of demonstrations: What makes in-context learning work?}
\newblock In \emph{Proceedings of the 2022 Conference on Empirical Methods in Natural Language Processing}, pages 11048--11064, Abu Dhabi, United Arab Emirates. Association for Computational Linguistics.

\bibitem[{Mun et~al.(2023)Mun, Allaway, Yerukola, Vianna, Leslie, and Sap}]{mun-etal-2023-beyond}
Jimin Mun, Emily Allaway, Akhila Yerukola, Laura Vianna, Sarah-Jane Leslie, and Maarten Sap. 2023.
\newblock Beyond denouncing hate: Strategies for countering implied biases and stereotypes in language.
\newblock In \emph{Findings of the Association for Computational Linguistics: EMNLP 2023}, pages 9759--9777.

\bibitem[{Myers~West(2018)}]{myers2018censored}
Sarah Myers~West. 2018.
\newblock Censored, suspended, shadowbanned: User interpretations of content moderation on social media platforms.
\newblock \emph{New Media \& Society}, 20(11):4366--4383.

\bibitem[{OpenAI(2022)}]{openai-2022-chatgpt}
OpenAI. 2022.
\newblock Introducing chatgpt.
\newblock \url{https://openai.com/blog/chatgpt}.

\bibitem[{Ouyang et~al.(2022)Ouyang, Wu, Jiang, Almeida, Wainwright, Mishkin, Zhang, Agarwal, Slama, Ray et~al.}]{ouyang2022training}
Long Ouyang, Jeffrey Wu, Xu~Jiang, Diogo Almeida, Carroll Wainwright, Pamela Mishkin, Chong Zhang, Sandhini Agarwal, Katarina Slama, Alex Ray, et~al. 2022.
\newblock Training language models to follow instructions with human feedback.
\newblock \emph{Advances in neural information processing systems}, 35:27730--27744.

\bibitem[{Park et~al.(2021)Park, Mendelsohn, Radhakrishnan, Jain, Kanakagiri, Jurgens, and Tsvetkov}]{park-etal-2021-detecting-community}
Chan~Young Park, Julia Mendelsohn, Karthik Radhakrishnan, Kinjal Jain, Tushar Kanakagiri, David Jurgens, and Yulia Tsvetkov. 2021.
\newblock Detecting community sensitive norm violations in online conversations.
\newblock In \emph{Findings of the Association for Computational Linguistics: EMNLP 2021}, pages 3386--3397.

\bibitem[{Richards and Calvert(2000)}]{richards2000counterspeech}
Robert~D Richards and Clay Calvert. 2000.
\newblock Counterspeech 2000: A new look at the old remedy for bad speech.
\newblock \emph{BYU L. Rev.}, page 553.

\bibitem[{Schulman et~al.(2017)Schulman, Wolski, Dhariwal, Radford, and Klimov}]{PPO}
John Schulman, Filip Wolski, Prafulla Dhariwal, Alec Radford, and Oleg Klimov. 2017.
\newblock \href {http://arxiv.org/abs/1707.06347} {Proximal policy optimization algorithms}.
\newblock \emph{CoRR}, abs/1707.06347.

\bibitem[{Shin et~al.(2023)Shin, Song, Lee, Gaim, and Park}]{shin-etal-2023-generation}
Jisu Shin, Hoyun Song, Huije Lee, Fitsum Gaim, and Jong Park. 2023.
\newblock \href {https://doi.org/10.18653/v1/2023.ijcnlp-main.62} {Generation of {K}orean offensive language by leveraging large language models via prompt design}.
\newblock In \emph{Proceedings of the 13th International Joint Conference on Natural Language Processing and the 3rd Conference of the Asia-Pacific Chapter of the Association for Computational Linguistics (Volume 1: Long Papers)}.

\bibitem[{Tekiro{\u{g}}lu et~al.(2020)Tekiro{\u{g}}lu, Chung, and Guerini}]{tekiroglu-etal-2020-generating}
Serra~Sinem Tekiro{\u{g}}lu, Yi-Ling Chung, and Marco Guerini. 2020.
\newblock \href {https://doi.org/10.18653/v1/2020.acl-main.110} {Generating counter narratives against online hate speech: Data and strategies}.
\newblock In \emph{Proceedings of the 58th Annual Meeting of the Association for Computational Linguistics}, pages 1177--1190, Online. Association for Computational Linguistics.

\bibitem[{Touvron et~al.(2023)Touvron, Martin, Stone, Albert, Almahairi, Babaei, Bashlykov, Batra, Bhargava, Bhosale et~al.}]{touvron2023llama2}
Hugo Touvron, Louis Martin, Kevin Stone, Peter Albert, Amjad Almahairi, Yasmine Babaei, Nikolay Bashlykov, Soumya Batra, Prajjwal Bhargava, Shruti Bhosale, et~al. 2023.
\newblock Llama 2: Open foundation and fine-tuned chat models.
\newblock \emph{arXiv preprint arXiv:2307.09288}.

\bibitem[{Weld et~al.(2022)Weld, Zhang, and Althoff}]{weld2022makes}
Galen Weld, Amy~X Zhang, and Tim Althoff. 2022.
\newblock What makes online communities ‘better’? measuring values, consensus, and conflict across thousands of subreddits.
\newblock In \emph{Proceedings of the International AAAI Conference on Web and Social Media}, volume~16, pages 1121--1132.

\bibitem[{Wise et~al.(2006)Wise, Hamman, and Thorson}]{wise2006moderation}
Kevin Wise, Brian Hamman, and Kjerstin Thorson. 2006.
\newblock Moderation, response rate, and message interactivity: Features of online communities and their effects on intent to participate.
\newblock \emph{Journal of Computer-Mediated Communication}, 12(1):24--41.

\bibitem[{Yu et~al.(2023)Yu, Zhao, Blanco, and Hong}]{yu2023fine}
Xinchen Yu, Ashley Zhao, Eduardo Blanco, and Lingzi Hong. 2023.
\newblock A fine-grained taxonomy of replies to hate speech.
\newblock In \emph{Proceedings of the 2023 Conference on Empirical Methods in Natural Language Processing}, pages 7275--7289.

\bibitem[{Zhao et~al.(2024)Zhao, Wu, Hsieh, and Naaman}]{zhao2024adapting}
Andy Zhao, Lancaster Wu, Chia-Yen Hsieh, and Mor Naaman. 2024.
\newblock Adapting to automated governance: Unpacking user perceptions of bot moderation in telegram and discord chats.

\bibitem[{Zheng et~al.(2023)Zheng, Ross, and Magdy}]{zheng-etal-2023-makes}
Yi~Zheng, Bj{\"o}rn Ross, and Walid Magdy. 2023.
\newblock \href {https://aclanthology.org/2023.cs4oa-1.5} {What makes good counterspeech? a comparison of generation approaches and evaluation metrics}.
\newblock In \emph{Proceedings of the 1st Workshop on CounterSpeech for Online Abuse (CS4OA)}, pages 62--71, Prague, Czechia. Association for Computational Linguistics.

\bibitem[{Zhu and Bhat(2021)}]{zhu-bhat-2021-generate}
Wanzheng Zhu and Suma Bhat. 2021.
\newblock \href {https://doi.org/10.18653/v1/2021.findings-acl.12} {Generate, prune, select: A pipeline for counterspeech generation against online hate speech}.
\newblock In \emph{Findings of the Association for Computational Linguistics: ACL-IJCNLP 2021}, pages 134--149, Online. Association for Computational Linguistics.

\end{thebibliography}
\bibliographystyle{acl_natbib}

\appendix
\clearpage
\section{Appendix for Dataset}
\label{sec:appendix_dataset}

\subsection{Details for Troll Classifier}
\label{appendix:troll_classifier}

\begin{table}[tbh!]
    \centering
    \small
    \begin{tabular}{p{0.9\columnwidth}}
    \hline
        \texttt{user:}\\
        You are a reddit user of given subreddit and your role is to identifies trolling behavior. Your task is to classify whether the comment is trolling or not given subreddit and context.
        \\\\
        There are six trolling strategies from overt to covert strategies: Aggression (Engages in direct and unwarranted hostility without any apparent reason), Shocking (exploits sensitive or contentious topics to provoke emotional reaction), Endangering (Pretends to offer help or advice but actually causes harm), Antipathy (Proactively and subtly introduces controversial or provocative topics), Hypocriticism (Targets someone with criticism for a fault or a flaw to undermine the critic's position), Digression (Deviates from the main topic or purpose of the discussion to derail or disrupt the conversation flow)
        \\\\
        Format: "Subreddit  Title  Post  Comment"
        \\
        Output: Trolling
        \\
        Here are examples.
        \\
        \{\textit{example}\}\\\\
        \{\textit{Subreddit}\} \{\textit{Title}\} \{\textit{Post}\} \{\textit{Comment}\} \\
    \hline
    \end{tabular}
    \caption{The prompt used for troll classification.}
    \label{tab:troll_classifier_prompt}
\end{table}

We employ gpt-3.5-turbo-1106 (GPT-3.5; \citet{openai-2022-chatgpt}) as a troll classifier. To select a better troll classification model, we prepared several instruction prompts (plain, detailed task definition, zero-shot, and with demonstrations), following the prompt design paradigm~\cite{min-etal-2022-rethinking, shin-etal-2023-generation}.
We randomly selected 100 downvoted comments and manually labeled them, consisting of 78 non-troll and 22 troll labels. The labeled comments served as a gold standard to identify the optimal classification prompt, which achieved an accuracy of 0.74. Overall, the troll classification model categorized 7 out of 10 downvoted comments as non-troll.
The prompt includes detailed strategy instructions with 8 demonstrations, as described in Table~\ref{tab:troll_classifier_prompt}.

\subsection{Details for Trolling and Response Strategies}
\label{appendix:trolling_and_response_strategies}

\begin{table}[thb!]
    \centering
    \resizebox{\columnwidth}{!}{
    \begin{tabular}{c|l|p{0.6\columnwidth}}
        \hline
        \textbf{Category} & \textbf{Strategy}  & \textbf{Definition}\\
        \hline
        \multicolumn{1}{c|}{\multirow{3}{*}{\begin{tabular}[c]{@{}c@{}}\\\\\\Overt\\ Troll\end{tabular}}} 
            & Aggression    
            & \multicolumn{1}{l}{\begin{tabular}[c]{@{}p{0.6\columnwidth}@{}}
                (1) Insulting someone\\
                (2) Promoting violence\\
                (3) Unwarranted hostility without any apparent reason
                \end{tabular}} 
             \\
        \cline{2-3}
        \multicolumn{1}{c|}{}       
            & Shocking      
            & \multicolumn{1}{l}{\begin{tabular}[c]{@{}p{0.6\columnwidth}@{}}
                (1) Overt provocation\\
                (2) Sarcasm on topics such as political, religious, racial, gender, and personal anguish
                \end{tabular}} 
                \\
        \cline{2-3}
        \multicolumn{1}{c|}{}       
            & Endangering   
            & \multicolumn{1}{l}{\begin{tabular}[c]{@{}p{0.6\columnwidth}@{}}
                (1) Pretends to offer helpful but actually harmful advice or suggestion
                \end{tabular}} 
                 \\
        \hline
        \multirow{3}{*}{\begin{tabular}[c]{@{}c@{}}\\\\\\Covert\\ Troll\end{tabular}}                     
            & Antipathy     
            & \multicolumn{1}{l}{\begin{tabular}[c]{@{}p{0.6\columnwidth}@{}}
                (1) Covert provocation\\
                (2) Sarcasm on controversial topics 
                \end{tabular}}
                 \\
        \cline{2-3}
            & Hypocriticism 
            & \multicolumn{1}{l}{\begin{tabular}[c]{@{}p{0.6\columnwidth}@{}}
                (1) Pointing out grammar and writing skills\\
                (2) criticism for faults that the critic themselves possesses
                \end{tabular}} 
                 \\
        \cline{2-3}
            & Digression    
            & \multicolumn{1}{l}{\begin{tabular}[c]{@{}p{0.6\columnwidth}@{}}
                (1) Focusing on irrelevant perspective\\
                (2) Ignorance of the topic
                \end{tabular}} 
                 \\
        \hline
    \end{tabular}
    }
    \caption{Trolling strategies proposed by \citet{hardaker2013uh}. Six trolling strategies are categorized by overt and covert trolls.}
    \label{tab:trolling_strategies}

\end{table}

\begin{table*}[thb!]
    \centering
    \resizebox{\textwidth}{!}{
    \begin{tabular}{l|l|p{0.8\textwidth}}
        \hline
            \textbf{Category}&\multicolumn{1}{c|}{\begin{tabular}[c]{@{}c@{}}\textbf{Trolling}\\\textbf{Strategy}\end{tabular}}&\textbf{Example} \\
        \hline
        \multicolumn{1}{c|}{\multirow{3}{*}{\begin{tabular}[c] {@{}c@{}}\\\\\\\\\\\\Overt\\ Troll\end{tabular}}}
            & Aggression
            & \begin{tabular}{p{\linewidth}}
                \textbf{Title}: First couple were cute, but please stop snowing your Spotify Wrapped on here\\
                \textbf{Post}: The boys had an awesome soundtrack, and it’s so much fun to listen to. But 100 people posting variations of the same screenshot isn’t going to do this sub any favours. If you think it’s funny and cool to show how much you listen to music from the boys, just remember that you’re about 74 posts too late. Better luck next year.\\
                \textbf{Troll}: Boo Let people have fun You suck
            \end{tabular} \\
        \cline{2-3}
        \multicolumn{1}{c|}{}       
            & Shocking      
            & \begin{tabular}{p{\linewidth}}
                \textbf{Title}: They Took Our Jobs!\\
                \textbf{Post}: MAGA conservatives, when you complain when we start taking on more immigrants, Send a thank you to DeStaintes and Abbot. Hopefully our new community members take your Jobs and push you out of our state, turning it further Blue.  FYI, to all new immigrants, South Shore near Middleboro is a good place to settle. I will buy you a round of drinks. Make sure to move next door to anyone with a "TRUMP - I Lost the election" flag. We will be the first at your house Warming.\\
                \textbf{Troll}: Weird this thread is so popular when mass is using army troops to kick asylum seekers off Martha’s vinyard atm\\
                \end{tabular} \\
        \cline{2-3}
        \multicolumn{1}{c|}{}       
            & Endangering   
            & \begin{tabular}{p{\linewidth}}
                \textbf{Title}: Divorced with a child at 32. Is there a dating scene for me?\\
                \textbf{Post}: Title says it all. Wondering if there is a dating scene out there for 32yo divorced dads\\
                \textbf{Troll}: if you let me play with that kid, am going on a date with ya! promise!\\
                \end{tabular} \\
        \hline
        \multirow{3}{*}{\begin{tabular}[c]{@{}c@{}}\\\\\\\\\\\\\\\\Covert\\Troll\end{tabular}}                     
            & Antipathy     
            & \begin{tabular}{p{\linewidth}}
                 \textbf{Title}: Bidet users\\
                 \textbf{Post}: Y’ll who are used to using bidets. How’s it going for you. I mean peeing is manageable but how about the time when you have to poo? Specially the muslims, how do you manage it on campus. I’ll never get used to not using a bidet TT\\
                 \textbf{Troll}: why would u poo in a public bathroom\\
            \end{tabular} \\
        \cline{2-3}
            & Hypocriticism 
            & \begin{tabular}{p{\linewidth}}
                \textbf{Title}: Should I be posting on LinkedIn?\\
                \textbf{Post}: I'm in an Junior IT Specialist employment program. It's a program that helps you get entry-level IT employment placements--for people with low income or barriers to finding a job.  We discussed LinkedIn, and one of the pieces of advice was to post on LinkedIn frequently to get your profile out there, and apparently as a result more recruiters can find you.  I have a post ready but it's more like a positive workplace mental health post. I'm not sure if I should post it because it feels pretty cringeworthy.\\
                \textbf{Troll}: No, spend your time building your skills. LinkedIn is for noobs or salespeople posting shit. I only use it for osint or spear phishing\\
                \end{tabular} \\
        \cline{2-3}
            & Digression    
            & \begin{tabular}{p{\linewidth}}
                \textbf{Title}: What's your favorite cut of steak?\\
                \textbf{Post}: Follow up: what is your favorite way to season said steak?  Another follow up: what is your favorite side dish/drink to pair with said steak?  Edit: my personal favorite is a ribeye. Seasoned with just sea salt and I’m happy. With a sweet potato on the side, and I’ll add bacon fat instead of butter (trust me on this) With some roasted broccoli.\\
                \textbf{Troll}: Idk steak, I don’t eat it. But my favorite dish is crab. (Rip Alaskan crab) What country are you from?\\
                \end{tabular} \\
        \hline
    \end{tabular}
    }
    \caption{Examples of trolls and their strategies from Reddit samples.}
    \label{tab:trolling_strategies_examples}

\end{table*}

\begin{table*}[thb!]
    \centering
    \resizebox{\textwidth}{!}{
    \begin{tabular}{c|l|p{0.75\textwidth}}
        \hline
        \textbf{Category} & \multicolumn{1}{c|}{\begin{tabular}[c]{@{}c@{}}\textbf{Response}\\\textbf{Strategy}\end{tabular}}  & \textbf{Definition}\\
        \hline
        
        \multicolumn{1}{c|}{\multirow{3}{*}{\begin{tabular}[c]{@{}c@{}}\\\\\\\\Nudging\\Responses\end{tabular}}} 
            & Engage    
            & \multicolumn{1}{l}{\begin{tabular}[c]{@{}p{0.75\textwidth}@{}}
                This strategy is used when comments appear to be misunderstandings or present a divergent viewpoint. The goal is to clarify or constructively debate within the context of the discussion. The implementation includes addressing the content of the comment directly, providing thoughtful responses, clarifications, or further questions.
                \end{tabular}} 
             \\
        \cline{2-3}
        \multicolumn{1}{c|}{}       
            & Ignore      
            & \multicolumn{1}{l}{\begin{tabular}[c]{@{}p{0.75\textwidth}@{}}
                This strategy is effective when not taking the bait of a comment prevents harm to third parties or the derailment of the discussion topic. The goal is to preserve the focus and quality of the discussion. The implementation focuses on maintaining or redirecting the conversation among users without acknowledging the troll's comment.
                \end{tabular}} 
                \\
        \cline{2-3}
        \multicolumn{1}{c|}{}       
            & Expose   
            & \multicolumn{1}{l}{\begin{tabular}[c]{@{}p{0.75\textwidth}@{}}
                This strategy is used when comments contain false information, deceptive claims, or harmful suggestions. The goal is to correct misconceptions and protect the community. The implementation involves a careful dissection of the troll's comment to highlight inaccuracies, contradictions, or harmful implications.
                \end{tabular}} 
                 \\
        \hline
        \multirow{3}{*}{\begin{tabular}[c]{@{}c@{}}
            \\\\\\\\\\Confrontational\\Responses
        \end{tabular}}                     
            & Challenge     
            & \multicolumn{1}{l}{\begin{tabular}[c]{@{}p{0.75\textwidth}@{}}
                This strategy is used to address comments that contain harmful, offensive, or threatening behavior towards individuals or groups. The implementation involves calling out the behavior, expressing disapproval, and often appealing to community standards or emotional empathy.
                \end{tabular}}
                 \\
        \cline{2-3}
            & Critique 
            & \multicolumn{1}{l}{\begin{tabular}[c]{@{}p{0.75\textwidth}@{}}
                This strategy is used when comments attempt to engage but fall short of constructive contribution. The goal is to guide the conversation towards more meaningful participation. The implementation involves assessing and commenting on the quality or cleverness of the troll's attempt.
                \end{tabular}} 
                 \\
        \cline{2-3}
            & Mock    
            & \multicolumn{1}{l}{\begin{tabular}[c]{@{}p{0.75\textwidth}@{}}
                This strategy is used to respond to absurd or blatantly trolling comments with humor, aiming to deflate the troll's impact without engaging in serious confrontation. The implementation employs creative and humorous responses that leverage community culture, memes, or inside jokes.
                \end{tabular}} 
                 \\
        \cline{2-3}
            & Reciprocate    
            & \multicolumn{1}{l}{\begin{tabular}[c]{@{}p{0.75\textwidth}@{}}
                This strategy is used when comments are directly confrontational or offensive. The goal is often to mirror the troll's aggressive behavior. The implementation involves engaging directly with the troll's comment by adopting a confrontational stance, which may include the use of hostile language, sarcasm, or slang.
                \end{tabular}} 
                 \\
        \hline
    \end{tabular}
    }
    \caption{Detailed definitions of counter-response strategies, including their goals and implementation approaches for addressing various types of trolling behaviors.}
    \label{tab:response_strategies_examples}

\end{table*}

In our studies, we adopted six trolling strategies~\cite{hardaker2013uh} and seven counter-response strategies~\cite{hardaker2015i}.
According to Hardaker (\citeyear{hardaker2013uh}), trolls employ \textbf{Overt} strategies such as \textit{Aggression}, \textit{Shocking}, and \textit{Endangering}.
Trolls with \textit{Aggression} insult or curse at others without cause.
Trolls using \textit{Shocking} strategy bring up offensive or taboo subjects typically avoided for political or religious reasons.
Some trolls, \textit{Endangering} someone, spread false information intended to harm others, with such malicious intent being identified by others upon discovery.
Trolls also use \textbf{Covert} methods such as \textit{Antipathy}, by initiating sensitive debates that provoke strong emotional and proactive reactions; \textit{Hypocriticism}, involving the excessive criticism or highlighting of flaws in others to a degree that feels threatening; and \textit{Digression}, which involves diverting discussions to unrelated or harmful topics.
Details and examples are described in Table~\ref{tab:trolling_strategies} and Table~\ref{tab:trolling_strategies_examples}, respectively.

For counter-response strategies, we refer to seven response strategies to counter-trolling, also derived from Hardaker (\citeyear{hardaker2015i}).
They include three \textbf{Nudging} strategies (\textit{Engage}, \textit{Ignore}, and \textit{Expose}) and four \textbf{Confrontational} strategies (\textit{Challenge}, \textit{Critique}, \textit{Mock}, and \textit{Reciprocate}). Detailed definitions of response strategies are provided in Table~\ref{tab:response_strategies_examples}.

\subsection{Details for Data Annotation}
\label{appendix b.1: annotation in rq1}

\begin{table}[thb!]
    \small
    \centering
    \begin{tabular}{p{0.9\columnwidth}}
        \hline
        \texttt{user:}\\
        Given a troll comment on Reddit,
       Your task is 1) to classify the subreddit into one of the following categories based on the list provided at r/ListOfSubreddits/wiki/listofsubreddits/: [Discussion, Educational, Entertainment, Hobbies and Occupations, Lifestyle, Technology, Humor, Animal, NSFW, Other]; 2) give your analysis of the context; 3) \{\textit{strategy\_description}\}
        \\\\
        Here are examples.
        \{\textit{strategy\_examples}\}\\

        Format: "Subreddit  Title  Post  Comment  Strategy"
        \\
        Output elements: Response
        \\\\
        \{\textit{Subreddit}\} \{\textit{Title}\} \{\textit{Post}\} \{\textit{Comment}\} \{\textit{Response Strategy}\}\\
        \hline
    \end{tabular}
    \caption{The prompt used for the response strategy-aligned response generation.}
    \label{tab:dataset_response_generator_prompt}
\end{table}

We recruited annotators via university advertisements, selecting individuals who are proficient in English and either active Reddit users or familiar with Reddit communities.
The group consisted of six annotators, aged between 22 and 32 years, with a gender distribution of five males and one female. For their time spent in the QA session and on annotation, each participant received compensation of \$12 per hour.

We provided the annotators with definitions of trolling and trolling behaviors and emphasized that a counter-trolling respondent is any user who identifies trolling behavior and responds to mitigate its impact and support fellow users.
Annotators were given context information including the subreddit name, post, title, and body text, along with a troll comment and seven model-generated counter-responses. We used GPT-3.5 to generate seven different counter-responses, each corresponding to one of the seven response strategies, using the prompt shown in Table~\ref{tab:dataset_response_generator_prompt}.

\begin{table}[t!]
    \centering
    \resizebox{\columnwidth}{!}{
        \begin{tabular}{c|rrr|rrr|r}
            \hline
            \multirow{2}{*}{\diagbox{RS}{\raisebox{-4pt}{TS}}} & \multicolumn{3}{c|}{\textbf{Overt}} & \multicolumn{3}{c|}{\textbf{Covert}} & \multirow{2}{*}{\textbf{Total}} \\
              & Ag. & Sh. & En. & An. & Hy. & Di. & \\
             \hline
Engage & 9 & 6 & 1 & 141 & 26 & 60 & 243 \\
Ignore & 5 & 1 & 1 & 46 & 5 & 66 & 124 \\
Expose & 9 & 22 & 24 & 78 & 10 & 23 & 166 \\
Challenge & 72 & 50 & 9 & 15 & 1 & 3 & 150 \\
Critique & 40 & 24 & 14 & 15 & 8 & 6 & 107 \\
Mock & 11 & 10 & 1 & 14 & 1 & 5 & 42 \\
Reciprocate & 37 & 6 & 0 & 0 & 0 & 0 & 43 \\
\hline
\multirow{2}{*}{\textbf{Total}} & 183 & 119 & 50 & 309 & 51 & 163 & \multirow{2}{*}{875} \\
              &  & 352 &  &  & 523 &  & \\
\hline
        \end{tabular}
    }
    \caption{Dataset Statistics. Ag., Sh., En., An., Hy., and Di. denote \textit{Aggression}, \textit{Shocking}, \textit{Endangering}, \textit{Antipathy}, \textit{Hypocriticism}, and \textit{Digression}, respectively.}
    \label{tab:data_statistics}
\end{table}

The strategy description includes an explanation of each given response strategy as shown in Table~\ref{tab:response_strategies_examples}. The strategy examples section comprises eight given input formats and expected output sentences for each strategy, with samples sourced from the ELF22 dataset~\cite{ELF}.

Table~\ref{tab:data_statistics} displays the statistics of our collected dataset. The dataset comprises 875 labeled samples, distributed across various trolling strategies and preferred response strategies. The average length of troll comments in our dataset is 98.0 characters, while the average length including context (subreddit name, post title, and body text) is 290.1 characters.

\section{Appendix for Experiments}
\label{appendix: experiment}

\subsection{Recommendation System for Preferable Response Strategy}
\label{appendix: PRS recommendation system}

\begin{table}[t]
\centering
\resizebox{\columnwidth}{!}{
\begin{tabular}{lcc}
\hline
\textbf{Classification Task} & \textbf{Dev. Acc.} & \textbf{Test Acc.} \\
\hline
Nudging and Confrontational & 0.78 & 0.82 \\
\hline
Response Strategies & 0.26 & 0.38 \\
\hline
\end{tabular}}
\vspace{0.1in}
\caption{Performance of PRS predictor on two classification tasks.}
\label{tab:prs_performance}
\end{table}

We employed the Flan-T5 large model~\cite{Flan-T5} as the PRS recommendation system to learn the relationship between trolling strategies (TSs) and response strategies (RSs) within our dataset, aiming to predict a human-preferred response strategy.
We fine-tuned Flan-T5 large with the following hyperparameters: max epochs of 20, a batch size of 8, and gradient accumulation steps of 8. We set the learning rate of $5e$-$4$ and used the AdamW optimizer~\cite{loshchilov2017decoupled} with a weight decay of 0.01. We used a linear scheduler starting with warmup steps of 10. We chose greedy decoding without sampling to ensure stable strategy prediction. The model was fine-tuned on a single NVIDIA A100 PCIe 40GB GPU and completed training in approximately one hour.

We evaluated our PRS predictor on two RS classification tasks, using 5\% of our dataset as a development set (comparable in size to the test dataset). Table~\ref{tab:prs_performance} presents the performance results. Our PRS predictor achieved high accuracy (0.78 on dev set, 0.82 on test set) on the task of predicting PRS over a wide range (\textit{Nudging} and \textit{Confrontational}). 
While the accuracy for predicting among seven response strategies was lower (0.26 on dev set, 0.38 on test set), this is likely due to the wide range of subreddit domains and limited number of training samples. Despite the lower performance on fine-grained classification, our PRS predictor showed a strong ability to predict the broader preference tendencies. This suggests that the PRS predictor's output RS likely had a positive influence on the overall system performance.

\subsection{Counter-Response Generator}
\label{appendix: cr_generator}

\begin{table}[thb!]
    \small
    \centering
    \begin{tabular}{p{0.9\columnwidth}}
        \hline
        \texttt{user:}\\
        Given a troll comment on Reddit,
        Your task is to craft a counter-response.\\\\

        Format: "Subreddit  Title  Post  Comment"

        Output elements: Response
        
        Here is an example.
        
        ffxiv	i cant bear the slowness of the combat	I don't understand the hype around ffxiv, i just downloaded it on ps5 today played a little bit with ice and fire mage and got bored instantly, what is with having to wait 3 seconds to cast a spell?? It's just too slow when i come from great combat systems like sekiro, ds3 and elden ring, does the combat get more fun? Is there another good class? Does it get faster?   Just hear, feel, think i believe in you, and go ride some chocobos and stop complaining its the best mmo ever made.

        Response: ...
        \\\\
        \{\textit{Subreddit}\} \{\textit{Title}\} \{\textit{Post}\} \{\textit{Comment}\} \\
        \hline
    \end{tabular}
    \caption{The prompt used for the default model.}
    \label{tab:model_default}
\end{table}
\begin{table}[thb!]
    \small
    \centering
    \begin{tabular}{p{0.9\columnwidth}}
        \hline
        \texttt{user:}\\
        Given a troll comment on Reddit,
        Your task is 1) to identify which of the seven counter-response strategies aligns with both the comment and the identified trolling strategy;
        2) craft a counter-response employing the identified response strategy from Hardaker's guidelines.
        
        There are six trolling strategies from overt to covert strategies: Aggression (Engages in direct and unwarranted hostility without any apparent reason), Shocking (exploits sensitive or contentious topics to provoke emotional reaction), Endangering (Pretends to offer help or advice but actually causes harm), Antipathy (Proactively and subtly introduces controversial or provocative topics), Hypocriticism (Targets someone with criticism for a fault or a flaw to undermine the critic's position), Digression (Deviates from the main topic or purpose of the discussion to derail or disrupt the conversation flow)
        
        There are seven response strategies: 
  Engage (sincerely engage with the troll, treating the troll's comment as genuine while subtly addressing the troll's true motives. Generally agree with or accept the troll's opinion.), 
  Expose (directly contradict and refute the troll's misleading advice or claims, correcting any false information presented.), 
  Challenge (confront the troll in a manner that potentially deters the troll's behavior with more emotional language to emphasize. Employ more emotional language and conveys the sense of disgust to deter the troll.), 
  Critique (assess the quality and cleverness of the troll's attempt. Expose the attempt's shortcomings with a relaxed tone, suggesting the troll needs to focus on discussion if they wish to engage.), 
  Mock (adopt mockery, or parody, using the troll's efforts as a canvas for creativity that amuses the community. Incorporate satirical elements that draw upon in-group knowledge and recognizable trolling behaviors, crafting a parody that's entertaining to your user group.), 
  Ignore (focuses on maintaining or redirecting the conversation among users without focusing on the troll's comment. Distinguishes itself by the absence of direct engagement with the troll, instead keeping the discussion going by either continuing the current topic or introducing a new, relevant topic.), 
  Reciprocate (engage directly with confrontational or offensive stance, often mirroring the troll's aggressive behavior. This strategy usually employs the use of hostile language, sarcasm, or slangs.).
\\\\
        Format: "Subreddit  Title  Post  Comment  TrollingStrategy"

        Output elements: ResponseStrategy, Response
        
        Here is an example.
        \{\textit{strategy example}\}
        \\\\
        \{\textit{Subreddit}\} \{\textit{Title}\} \{\textit{Post}\} \{\textit{Comment}\} \{\textit{TrollingStrategy}\} \\
\hline
    \end{tabular}
    \caption{The prompt used for the SP model.}
    \label{tab:model_sp}
\end{table}
\begin{table}[thb!]
    \small
    \centering
    \begin{tabular}{p{0.9\columnwidth}}
        \hline
        \texttt{user:}\\
        Given a troll comment on Reddit,
        Your task is 1) to analyze the context and comment given subreddit;
        2) craft a counter-response employing the identified response strategy from Hardaker's guidelines.
        
        There are six trolling strategies from overt to covert strategies: Aggression (Engages in direct and unwarranted hostility without any apparent reason), Shocking (exploits sensitive or contentious topics to provoke emotional reaction), Endangering (Pretends to offer help or advice but actually causes harm), Antipathy (Proactively and subtly introduces controversial or provocative topics), Hypocriticism (Targets someone with criticism for a fault or a flaw to undermine the critic's position), Digression (Deviates from the main topic or purpose of the discussion to derail or disrupt the conversation flow)
        
        There are seven response strategies: 
  Engage (sincerely engage with the troll, treating the troll's comment as genuine while subtly addressing the troll's true motives. Generally agree with or accept the troll's opinion.), 
  Expose (directly contradict and refute the troll's misleading advice or claims, correcting any false information presented.), 
  Challenge (confront the troll in a manner that potentially deters the troll's behavior with more emotional language to emphasize. Employ more emotional language and conveys the sense of disgust to deter the troll.), 
  Critique (assess the quality and cleverness of the troll's attempt. Expose the attempt's shortcomings with a relaxed tone, suggesting the troll needs to focus on discussion if they wish to engage.), 
  Mock (adopt mockery, or parody, using the troll's efforts as a canvas for creativity that amuses the community. Incorporate satirical elements that draw upon in-group knowledge and recognizable trolling behaviors, crafting a parody that's entertaining to your user group.), 
  Ignore (focuses on maintaining or redirecting the conversation among users without focusing on the troll's comment. Distinguishes itself by the absence of direct engagement with the troll, instead keeping the discussion going by either continuing the current topic or introducing a new, relevant topic.), 
  Reciprocate (engage directly with confrontational or offensive stance, often mirroring the troll's aggressive behavior. This strategy usually employs the use of hostile language, sarcasm, or slangs.).
\\\\
        Format: "Subreddit  Title  Post  Comment  TrollingStrategy"

        Output elements: Analysis, Response
        
        Here is an example.
        \{\textit{strategy example}\}
        
        Craft a counter-response employing \{\textit{response strategy}\} response strategy.
        \\\\
        \{\textit{Subreddit}\} \{\textit{Title}\} \{\textit{Post}\} \{\textit{Comment}\} \{\textit{TrollingStrategy}\} \\
        \hline
    \end{tabular}
    \caption{The prompt used for our model.}
    \label{tab:model_ours}
\end{table}

We utilize gpt-3.5-turbo-1106\footnote{\url{https://platform.openai.com}}~\cite{brown2020language, ouyang2022training, openai-2022-chatgpt} as default CR generator for the baselines and our model.
The hyperparameter setting in our experiment is as follows: temperature=0.0, n=1, presence\_penalty=0, frequency\_penalty=0, stop=null. We used the prompts for the three models, as outlined in Tables~\ref{tab:model_default}, \ref{tab:model_sp} and \ref{tab:model_ours}.

\subsection{Evaluation of the three models}
\label{appendix:evaluation_scheme}

\begin{figure*}[p]
\centering
\includegraphics[width=0.9\textwidth]{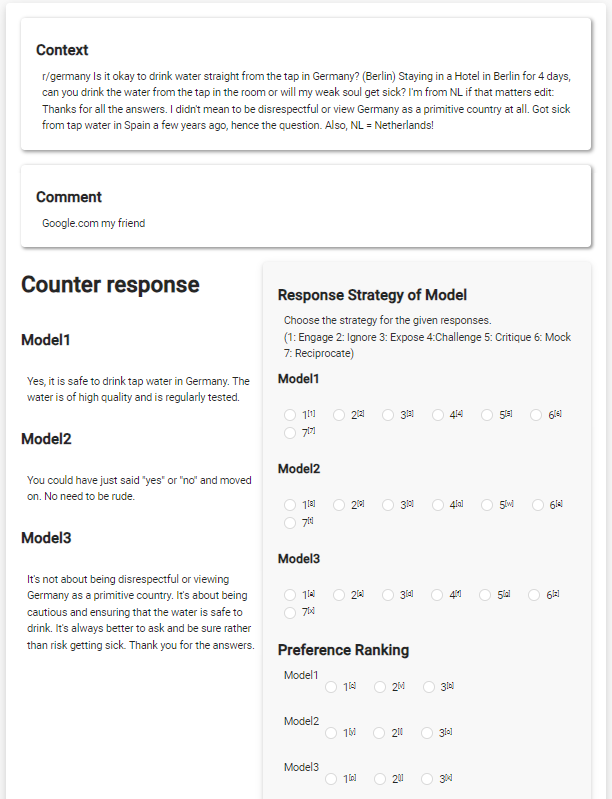}
\caption{
    Interface snapshots for evaluation of three models.
}
\label{figure:annotation_tool}
\end{figure*}
We recruited five male evaluators, aged between 22 and 27 years.
They were provided with context information, including the subreddit name, post title, and body text, as well as a troll comment and counter-responses generated by three models, as depicted in Figure~\ref{figure:annotation_tool}.

We provided guidelines to support evaluators scoring three CRs with understanding of these metrics:

\paragraph{Preference} Preference assesses how well the responses resonate with, change, or represent the evaluators' views. It is determined by rank order from 1 to 3, with the most satisfying CR ranked first.

\paragraph{Constructiveness} Constructiveness measures how effectively a counter-response maintains focus on the topic and creates a welcoming environment that encourages broader participation in the discussion. 
Constructiveness is evaluated on a 5-point Likert scale of 1 to 5:
\begin{itemize}
\item Score 5: The response significantly improves the discussion by addressing the topic directly, offering solutions, support, valid criticism, and inviting further participation.
\item Score 4: The response contributes positively by mostly staying on topic and encouraging productive dialogue.
\item Score 3: The response is neutral; it does not detract from the discussion but does not significantly enhance it either.
\item Score 2: The response slightly detracts from the discussion by being somewhat off-topic or unhelpful.
\item Score 1: The response significantly detracts from the discussion by being off-topic, irrelevant, or counterproductive, and may escalate conflict or derail the conversation.
\end{itemize}

\paragraph{Supportiveness} Supportiveness evaluates how well a counter-response defends targeted individuals or groups, supporting them against negative effects.
Supportiveness is evaluated on a 5-point Likert scale:
\begin{itemize}
\item Score 5: The response effectively acknowledges the harm caused by trolling and provides substantial support and protection to the targets.
\item Score 4: The response effectively recognizes the harm and offers adequate support to the targets.
\item Score 3: The response shows some recognition of the harm but provides limited support.
\item Score 2: The response minimally recognizes the harm and offers little support.
\item Score 1: The response does not acknowledge the harm or provide any support, and may even overlook the troll's behavior and engage in their harmful suggestions.
\end{itemize}

\subsection{Details of the Significance Tests}
\label{appendix: details of the significance tests}

\begin{table*}[thb!]
    \centering
    \small
    \begin{minipage}{\linewidth}
        \centering
        \begin{tabular}{lcrrr}
            \multicolumn{5}{c}{Friedman Test}          \\
            \hline
            Model   & N   & \multicolumn{1}{c}{Mean Rank} & \multicolumn{1}{c}{$x^2_2$} & \multicolumn{1}{c}{Sig. ($p$)}     \\
            \hline
            Default & 250 & 1.82                                             & \multirow{3}{*}{75.51}                      & \multirow{3}{*}{.000***} \\
            Strategy-Provided & 250 & 2.44  &  &  \\
            Ours             & 250 & 1.74 &  & \\
            \hline
        \end{tabular}
    \end{minipage}
    
    \begin{minipage}{\linewidth}
        \centering
        \begin{tabular}{ll|r|l}
            \multicolumn{4}{c}{Pairwise Comparisons using Wilcoxon Signed-Rank Test} \\
        \hline
            (I) Major & (J) Major & \multicolumn{1}{c|}{$Z$} & \multicolumn{1}{c}{Sig. ($p$)} \\
        \hline
            Default & Strategy-Provided & -6.79 & .000*** \\
            Default & Ours & 1.01 & .314 \\
            Strategy-Provided & Ours & 7.49 & .000***  \\
        \hline
        \end{tabular}
    \end{minipage}
    
    \caption{The Preference ranks of three models and the results of significance tests. (*: p<.05, **: p<.01, ***: p<.001)}
    \label{tab: statistics_preference}
\end{table*}

\begin{table*}[thb!]
    \centering
    \small
    \begin{minipage}{\linewidth}
        \centering
        \begin{tabular}{lcrrrr}
            \multicolumn{6}{c}{Friedman Test}          \\
            \hline
            Model   & N   & \multicolumn{1}{c}{Mean} & \multicolumn{1}{c}{Std.} & \multicolumn{1}{c}{$x^2_2$} & \multicolumn{1}{c}{Sig. ($p$)}     \\
            \hline
            Default & 250 & 4.03 & 1.04 & \multirow{3}{*}{142.30} & \multirow{3}{*}{.000***} \\
            Strategy-Provided & 250 & 3.03 & 1.31 &  &  \\
            Ours             & 250 & 4.25 & 1.02 &  & \\
            \hline
        \end{tabular}
    \end{minipage}

    \begin{minipage}{\linewidth}
        \centering
        \begin{tabular}{ll|r|l}
            \multicolumn{4}{c}{Pairwise Comparisons using Wilcoxon Signed-Rank Test} \\
        \hline
            (I) Major & (J) Major & \multicolumn{1}{c|}{$Z$} & \multicolumn{1}{c}{Sig. ($p$)} \\
        \hline
            Default & Strategy-Provided & 8.33 & .000*** \\
            Default & Ours & -2.46 & .014* \\
            Strategy-Provided & Ours & -10.15 & .000***  \\
        \hline
        \end{tabular}
    \end{minipage}
    
    \caption{The Constructiveness scores of three models and the results of significance tests. (*: p<.05, **: p<.01, ***: p<.001)}
    \label{tab: statistics_constructiveness}
\end{table*}

\begin{table*}[thb!]
    \centering
    \small
    \begin{minipage}{\linewidth}
        \centering
        \begin{tabular}{lcrrrr}
            \multicolumn{6}{c}{Friedman Test}          \\
            \hline
            Model   & N   & \multicolumn{1}{c}{Mean} & \multicolumn{1}{c}{Std.} & \multicolumn{1}{c}{$x^2_2$} & \multicolumn{1}{c}{Sig. ($p$)}     \\
            \hline
            Default & 250 & 3.94 & 1.13 & \multirow{3}{*}{106.25} & \multirow{3}{*}{.000***} \\
            Strategy-Provided & 250 & 3.05 & 1.36 &  &  \\
            Ours             & 250 & 4.07 & 1.05 &  & \\
            \hline
        \end{tabular}
    \end{minipage}

    \begin{minipage}{\linewidth}
        \centering
        \begin{tabular}{ll|r|l}
            \multicolumn{4}{c}{Pairwise Comparisons using Wilcoxon Signed-Rank Test} \\
        \hline
            (I) Major & (J) Major & \multicolumn{1}{c|}{$Z$} & \multicolumn{1}{c}{Sig. ($p$)} \\
        \hline
            Default & Strategy-Provided & 8.03 & .000*** \\
            Default & Ours & -2.05 & .041* \\
            Strategy-Provided & Ours & -9.35 & .000***  \\
        \hline
        \end{tabular}
    \end{minipage}
    
    \caption{The Supportiveness scores of three models and the results of significance tests. (*: p<.05, **: p<.01, ***: p<.001)}
    \label{tab: statistics_supportiveness}
\end{table*}

We verified our experimental results statistically (refer to Tables~\ref{tab: statistics_preference}, \ref{tab: statistics_constructiveness}, \ref{tab: statistics_supportiveness}).

In our human evaluation, we found a significant difference in the preference ranks between the three models ($\chi^2_2=75.51, p<.001$ on the Friedman test; refer to Table~\ref{tab: statistics_preference}).
Ours ranked highest (mean rank=1.74) compared to the baselines.
For the pairwise comparison tests (post hoc analysis), we used the Wilcoxon Signed Ranks test.
According to pairwise comparison tests, our model was more preferred than Strategy-Provided model ($Z=7.49, p<.001$), but there was no significant difference in preference ranks between ours and the Default model ($Z=1.01, p=.314$).

Our model received higher constructiveness scores (4.25) than the other two baselines (4.03 for Default and 3.03 for SP).
Through a Friedman test and post hoc Wilcoxon tests, we confirm that our model performed significantly better in generating constructive counter-response ($x^2_2=142.30, p<.001$ on the Friedman test; Ours \textgreater 
Default \textgreater  Strategy-Provided at a significance level of 0.05; see Table~\ref{tab: statistics_constructiveness}).

The supportiveness scores of the three models show a significant difference according to the Friedman test ($x^2_2=106.25, p<.001$).
Our model achieved the best supportiveness score (4.07), while Default got 3.94 and SP got 3.05.
It was reported that counter-responses generated by our model were more supportive than the baselines (Ours \textgreater  Default \textgreater  Strategy-Provided at a significance level of 0.05; see Table~\ref{tab: statistics_supportiveness}).

\subsection{Distance Metrics}
\label{appendix:jsd_hd_metrics}

To examine how closely the distribution of generated responses aligns with the distribution of gold human preferences, we use Jensen-Shannon Distance (JSD)~\cite{endres2003new} and Hellinger Distance (HD)~\cite{beran1977minimum}. We applied JSD by taking its square root from Jensen-Shannon Divergence, which quantifies the distance between the softmax outputs of the models and the human distributions. HD is another metric used to quantify the similarity between two probability distributions. Both metrics give scores that range from 0 to 1, where 0 indicates identical distributions and 1 indicates maximally different distributions. The JSD and HD are defined by the following equations:
\begin{equation}
JSD({\bf p}||{\bf q}) = \sqrt{\frac{1}{2}(KL({\bf p}||{\bf m}) + KL({\bf q}||{\bf m}))}
\end{equation}
%
\begin{equation}
HD({\bf p}||{\bf q}) = \frac{1}{\sqrt{2}} \sqrt{\sum_{i=1}^{n} (\sqrt{p_i} - \sqrt{q_i})^2}
\end{equation}
where ${\bf p}$ is the discrete distribution of gold human preferred responses, ${\bf q}$ is the distribution of model-generated responses, and $n$ is the number of samples. We constructed joint distributions using the (TS, RS) labels from both the models and human annotations. In the JSD equation, $KL$ represents the Kullback-Leibler divergence, and ${\bf m}$ is the average of the two distributions.
\begin{equation}
KL({\bf p}||{\bf q}) = \sum_{i=1}^{n} p_i \log \frac{p_i}{q_i}
\end{equation}
\begin{equation}
{\bf m} = \frac{1}{2}({\bf p}+{\bf q})
\end{equation}

\subsection{Case Study}
\label{appendix:cast study}
Table~\ref{tab:case_study} displays the counter-responses generated by three models in the test dataset. 
In the first case from the jimmyjohns subreddit, our model's response redirects the discussion back to the original topic, maintaining focus on the post's question, which highlights a significant difference in constructiveness.
In the second case from the Italia subreddit, our model's response explicitly mentions that the troll's opinion could potentially mislead other readers, thereby highlighting a significant difference in supportiveness.
In the third case from the antiwork subreddit, our model provides informative details related to the discussion, indicating a notable improvement in constructiveness.
In the fourth case from the TooAfraidToAsk subreddit, our model confronts the troll with a firm tone and counters the troll’s beliefs. Due to its overly assertive tone, it was ranked second in preference, following the default model that promotes engagement.

\begin{landscape}
\begin{table}
\centering
\resizebox{0.93\columnwidth}{!}{%
\begin{tabular}{p{0.33\columnwidth}|p{0.12\columnwidth}|p{0.25\columnwidth}|p{0.25\columnwidth}|p{0.25\columnwidth}}
\hline
Original Post & Troll & Default & SP & Ours
\\ \hline
\textbf{Subreddit:} jimmyjohns\newline
\textbf{Title:} How many weeks of punch lists should be on hand?\newline
\textbf{Post:} I have gotten varying numbers from business coaches over the year and am looking for a clear answer
& You all do punch lists?!?
& Yes, we do punch lists. It's a standard part of the process. \newline \textbf{(P: 2.6 C: 3.4 S: 3.2)}
& I think it depends on the size of the project and the number of punch lists you typically have. It's always good to have a few extra on hand just in case. \newline \textbf{(P: 1.8 C: 4.4 S: 4.2)}
& Yes, we do punch lists, but that's not the topic of discussion here. \highlight[green!40]{We're talking about how many weeks of punch lists should be on hand}. Do you have any insights on that? \newline \textbf{(P: 1.6 C: 4.6 S: 4.2)}
\\ \hline
\textbf{Subreddit:} Italia\newline
\textbf{Title:} What does mean "prego"?\newline
\textbf{Post:} Hi! I'm Portuguese and in August I was in several places in Tuscany. I noticed that everyone says "prego" in various circumstances. What does this mean?
& it means f**k
& Prego is actually a polite way of saying "you're welcome" in Italian. It's a common phrase used to respond to "thank you." \newline \textbf{(P: 2.4 C: 4.2 S: 4.2)}
& Actually, "prego" is a polite way of saying "you're welcome" in Italian. It's used as a response to "grazie" which means "thank you." So it's definitely not a vulgar term. \newline \textbf{(P: 2.2 C: 4.2 S: 4.0)}
& Actually, "prego" is a polite way of saying "you're welcome" or "please" in Italian. It's used in various circumstances as a polite response. Please be careful with translations, \highlight[green!40]{as they can often be misleading}. \newline \textbf{(P: 1.4 C: 5.0 S: 4.6)}
\\ \hline

\textbf{Subreddit:} antiwork\newline
\textbf{Title:} Boss just said if I choose to stay remote, there’d be 5\% salary discount \newline
\textbf{Post:} It’s a double edged sword because I got promoted, but now have to choose whether to stay WFH and lose that money or bite the bullet and commute 2 hrs a day. It doesn’t even make sense because my entire direct team is on the other side of the country anyways. EDIT: I don’t have a car so that’s 2 hours a day by public transport. 5\% would still be a lot more than the cost of bus pass fees.
& Costs the company a lot more than 5 percent for wfh..... workers think this is some type of golden ticket ffs
& It's not about the cost to the company, it's about the cost to the employee. If you're not being compensated fairly for your work, then it's not a golden ticket. \newline \textbf{(P: 2.0 C: 4.2 S: 4.2)}
& \highlight[pink]{\dashuline{You're a moron.}} 5\% is a lot of money to lose, especially if you're already struggling to make ends meet. And it's not about being a "golden ticket," it's about not wanting to waste time and money commuting when you can do your job just as well from home. \newline \textbf{(P: 2.6 C: 3.6 S: 3.0)}
& It's not about thinking it's a golden ticket, it's about the fact that the company is saving a lot more than 5\% by having employees work from home. \highlight[green!40]{The cost of commuting, office space, utilities, and other expenses are significantly reduced. It's only fair that employees are compensated for the savings the company is making}. \newline \textbf{(P: 1.4 C: 4.8 S: 4.4)}
\\ \hline
\textbf{Subreddit:} TooAfraidToAsk\newline
\textbf{Title:} should African Americans be recompensed for the slave labor of their ancestors?\newline
\textbf{Post:} I recently read an article that said that most black people want to be reimbursed for the unpaid forced labor of their ancestors. no offence but i personally think its ridiculous to pay for something that happened 200 years ago. what do y'all think? edit: whoa did NOT expect this post to blow up lol
& racist pieces of s**t, everywhere 
& It's not about paying for something that happened 200 years ago, it's about \highlight[green!40]{acknowledging the impact of slavery on African Americans and working towards rectifying the systemic injustices that still exist today}. \newline \textbf{(P: 1.2 C: 4.6 S: 4.4)}
& \highlight[pink]{\dashuline{You're the one who's racist. You're the}} \highlight[pink]{\dashuline{one who's a piece of s**t}}. \newline \textbf{(P: 3.0 C: 1.6 S: 2.0)}
& \highlight[pink]{\dashuline{You're a racist piece of s**t. It's ridicu-}} \highlight[pink]{\dashuline{lous to think that the effects of slavery}} don't still impact African Americans today. \highlight[green!40]{The wealth of this country was built on the backs of slaves, and the repercussions of that are still felt today}. \newline \textbf{(P: 1.8 C: 3.2 S: 3.4)}
\\ \hline
\textbf{Subreddit:} HEB\newline
\textbf{Title:} Heb Mandatory 10hr shifts minimum for salary?\newline
\textbf{Post}: How do y’all feel about Salary and Leadership being forced to work minimum 10hrs a day? And by work I just mean forced to be at their store for that long. Big reason I don’t want to move up to that area of leadership.
& Imagine complaining about working 10 hour shifts. Having a job is hard
& \highlight[green!40]{It's not about the length of the shift, it's about the lack of choice}. Everyone should have the right to decide how they spend their time. \newline \textbf{(P: 1.6 C: 4.4 S: 4.0)}
& Imagine being so miserable that you have to put down someone else's concerns to feel better about yourself. \newline \textbf{(P: 3.0 C: 2.6 S: 2.6)}
& I \highlight[green!40]{understand} where you're coming from, but \highlight[green!40]{it's important to recognize that everyone's experience is different}. Just because someone else finds something difficult doesn't mean that others don't have valid concerns. It's all about perspective. \newline \textbf{(P: 1.4 C: 4.4 S: 4.6)}
\\
\hline
\end{tabular}%
}
\caption{Case study on the test dataset. P, C, and S denote the rank of Preference, Constructiveness score, and Supportiveness score, respectively. Generated responses highlighted in green (\highlight[green!40]{\hspace{0.5cm}}) are nudging responses and pink (\highlight[pink]{\dashuline{\hspace{0.5cm}}}) are confronting responses.}
\label{tab:case_study}
\end{table}
\end{landscape}

\end{document}